\documentclass[acmsmall]{acmart}

\usepackage{hyperref}
\usepackage{url}

\usepackage{amsmath, enumitem, array, bbm, graphicx, multirow, xcolor, booktabs, subcaption}

\setcopyright{none}
\renewcommand\footnotetextcopyrightpermission[1]{} 
\AtBeginDocument{%
  \providecommand\BibTeX{{%
    \normalfont B\kern-0.5em{\scshape i\kern-0.25em b}\kern-0.8em\TeX}}}

\begin{document}

\title{StructCoder: Structure-Aware Transformer for Code Generation}

\author{Sindhu Tipirneni}
\email{sindhut@vt.edu}
\affiliation{%
  \institution{Department of Computer Science, Virginia Tech, Arlington, VA}
  \country{USA}
}

\author{Ming Zhu}
\email{mingzhu@vt.edu}
\affiliation{%
  \institution{Department of Computer Science, Virginia Tech, Arlington, VA}
  \country{USA}
}

\author{Chandan K. Reddy}
\email{reddy@cs.vt.edu}
\affiliation{%
  \institution{Department of Computer Science, Virginia Tech, Arlington, VA}
  \country{USA}
}


\begin{abstract}
There has been a recent surge of interest in automating software engineering tasks using deep learning. This paper addresses the problem of code generation, where the goal is to generate target code given source code in a different language or a natural language description.
Most state-of-the-art deep learning models for code generation use training strategies primarily designed for natural language. However, understanding and generating code requires a more rigorous comprehension of the code syntax and semantics. With this motivation, we develop an encoder-decoder Transformer model where both the encoder and decoder are explicitly trained to recognize the syntax and data flow in the source and target codes, respectively.
We not only make the encoder structure-aware by leveraging the source code's syntax tree and data flow graph, but we also support the decoder in preserving the syntax and data flow of the target code by introducing two novel auxiliary tasks: AST (Abstract Syntax Tree) paths prediction and data flow prediction. To the best of our knowledge, this is the first work to introduce a structure-aware Transformer decoder that models both syntax and data flow to enhance the quality of generated code.
The proposed StructCoder model achieves state-of-the-art performance on code translation and text-to-code generation tasks in the CodeXGLUE benchmark, and improves over baselines of similar size on the APPS code generation benchmark. Our code is publicly available at \url{https://github.com/reddy-lab-code-research/StructCoder/}.
\end{abstract}

\ccsdesc[500]{Computing methodologies~Neural networks}
\ccsdesc[300]{Computing methodologies~Natural language processing}
\ccsdesc[300]{Software and its engineering~Automatic programming}

\keywords{deep learning, language models, code generation, transformer}


\maketitle

\section{Introduction}
Code generation is the problem of generating code in a specified target language given source code that is either imperfect or in a different language, or generating code from a natural language description.
In this paper, we consider the problem of generating target code given source code in a different language (code translation) or a natural language description (text-to-code generation).
Code translation has applications in migrating legacy codebases to contemporary programming languages and porting existing software to various other platforms \citep{transcoder, plbart, zhu2022multilingual}. 
Text-to-code generation models can potentially increase programmers' productivity by simplifying and speeding up the software development process, as developers often write code to solve a problem or implement logic that is stated in natural language\citep{plbart}. Transformer-based deep learning methods have recently gathered significant attention in this domain. However, these existing models do not effectively utilize the code structure, especially during the decoding of target code. To address this limitation, we propose StructCoder which models the syntax and data flow in both source and target codes with a structure-aware encoder and decoder.

Traditional code translation tools have been designed using hand-crafted rules based on the Abstract Syntax Tree (AST)\citep{transcoder}. One such popular tool is Babel\footnote{\url{https://github.com/babel/babel}} which converts modern JavaScript code to older versions for backward compatibility. Other notable source-to-source translators include c2go\footnote{https://github.com/elliotchance/c2go}, grumpy\footnote{https://github.com/google/grumpy}, TypeScript\footnote{https://github.com/microsoft/TypeScript}, etc. However, the design of such tools demands a lot of time and effort as it requires proficiency in both source and target languages \citep{zhu2022multilingual}. Moreover, such tools are specific to the particular programming language pairs they are designed for. Since the task of generating code from natural language text is more difficult than translation due to the inherent ambiguity in natural language, almost all text-to-code generation tools are AI-based.

Code generation bears a strong resemblance to natural language generation as both involve the creation of a sequence of words or tokens. 
Since natural language generation using deep learning has achieved great success in recent years, it is natural to exploit similar deep learning based approaches for code generation as well.
However, the code domain faces a unique set of challenges. Since the generated code is to be understood by a machine as opposed to a human, it is even more important for the generated code (compared to natural language) to adhere to a specific syntax. Moreover, since a minor change in code could alter its function, it is also critical to preserve the semantic information from the source code during translation. To generate syntactically correct code, some of the existing approaches for code generation leveraged the AST structure by learning to generate inorder traversal of AST \citep{li2017code}, learning to generate production rules for AST based on a grammar, encoding AST paths using RNNs \citep{alon2020structural}, and using AST-based attention \citep{li2017code, kim2021code} in sequence models. 
\citet{guo2020graphcodebert} hypothesize that Data Flow Graph (DFG), which contains more semantic information and is less complex than AST, is a more useful structure to learn code representations. They incorporate DFG into the Transformer encoder by appropriately masking the attention matrix. 
Our model, \textit{StructCoder}, consists of a Transformer encoder that incorporates both syntax and data flow of source code by embedding root-leaf paths in the AST and using a modified self-attention framework, called \textit{structure-aware self-attention}.

\textbf{Code generation heavily relies on the decoder to generate code that is syntactically correct while simultaneously preserving the semantics present in the input. Structcoder advances the state-of-the-art by incorporating a structure-aware Transformer decoder that is designed to preserve of syntax and semantics of the generated code.} None of the existing pretrained Transformer models constrain the generated code structure. In our work, we not only incorporate source AST and DFG into the encoder, but also drive the decoder to learn the target syntax and data flow by introducing novel AST and DFG related tasks. Particularly, we train the decoder to predict all the root-leaf paths in the target AST and also to predict the DFG edges.

Similar to pretrained language models \citep{devlin2018bert, liu2019roberta, radford2019language, yang2019xlnet}, pretrained code models using Transformer \citep{codebert,plbart,dobf,zugner2021language} have resulted in significant performance gains on code-related tasks. 
While some pretext tasks like Masked Language Modeling (MLM) and Replaced Token Detection (RTD) only pretrain the encoder, other pretext tasks like Denoising Autoencoding (DAE) and Back Translation (BT) jointly pretrain both the encoder and decoder. StructCoder falls in the latter category and is pretrained using a structure-based DAE task. Moreover, since the structure-based components introduced in this work can be added to any existing Transformer model, we may initialize most of the StructCoder weights using one of the pretrained code models to avoid pretraining from scratch which can be quite expensive.
The main contributions of this work are listed below:
\begin{enumerate}
    \item We develop a Transformer-based encoder-decoder model called StructCoder for code generation where both encoder and decoder are structure-aware. (a) The encoder incorporates AST's root-leaf path embeddings and a structure-aware self-attention framework to model source code structure. (b) The decoder is trained to recognize target syntax and data flow via two novel auxiliary tasks: AST paths prediction and data flow prediction.
    \item We pretrain StructCoder using a structure-based DAE objective where the input code as well as its AST and DFG  are partially corrupted and the model is trained to generate the original input code and also perform the auxiliary tasks.
    \item Our experiments demonstrate that the proposed model achieves state-of-the-art performance on the code translation and text-to-code generation tasks in the CodeXGLUE \citep{codexglue} benchmark, and outperforms similarly sized baselines on the APPS code generation benchmark.
\end{enumerate}

The subsequent sections of this paper are organized as follows. Section \ref{sec:related} discusses existing methods for modeling code structure and developing pretrained transformers for code. Section 3 provides a detailed description of our proposed methodology. In Section \ref{sec:experiments}, we present experimental results, comparing our model against the baselines on code translation and text-to-code generation datasets. We also conduct an ablation study and discuss more aspects StructCoder's performance. Finally, Section \ref{sec:conclusion} concludes the paper.
\section{Related Work} \label{sec:related}
\subsection{Leveraging Structure to Generate Code}
To leverage code structure in deep models, many approaches have utilized ASTs.
Some approaches modeled code completion as a language modeling task by ordering the code tokens using a depth-first traversal of AST. \citet{li2017code} used an LSTM appended with parent-child attention while \citet{alon2020structural} encoded each root-to-leaf path with an LSTM.
\citet{kim2021code} used the Transformer to encode the sequenced AST by encoding AST paths into self-attention.
For text-to-code generation,
\citet{rabinovich2017abstract} proposed a modular decoder to recursively generate target AST.
\citet{yin2017syntactic, brockschmidt2018generative, sun2020treegen} construct ASTs by generating production rules based on a grammar. 
\citet{jiang2021ast} proposed an LSTM decoder equipped with AST enhanced attention, to generate a sequence of production rules by attending to previously generated rules and one future rule. To go beyond the standard preorder traversal for AST node generation, \citet{jiang2021exploring} used a Reinforcement Learning framework for dynamically selecting the branch to expand at an intermediate AST node, and \citet{xie2021improving} used two separate models for preorder and breadth-first traversals that are jointly trained via mutual distillation. 
Unlike these methods, we keep the conventional Transformer decoder architecture intact and introduce auxiliary structure-related components on top of the decoder's final hidden representations, so that StructCoder is trained to preserve target code structure while not requiring the generation of such structures (AST/DFG) during inference. Building on top of the conventional Transformer architectures not only allows us to utilize existing pretrained models for better initialization but also makes the advances in the area of Transformers more easily applicable to our model. 




\begin{table*}[!h] \label{tab:related}
\caption{A summary of the recent pre-trained models for code generation. (Abbreviations: DFG: Data Flow Graph, MLM: Masked Language Modeling, DAE: Denoising Autoencoding, RTD: Replaced Token Detection, BT: Back Translation, EP: DFG Edge Prediction, NA: Alignment prediction between code tokens and DFG nodes, DOBF: Deobfuscation, IT: Identifier Tagging, MSP: Masked Span Prediction, MIP: Masked Identifier Prediction, MuST: Multilingual Snippet Translation.)}
\footnotesize
\centering
\begin{tabular}{l>{\centering\arraybackslash}p{2.3cm}>{\centering\arraybackslash}p{3cm}>{\centering\arraybackslash}p{1.7cm}>{\centering\arraybackslash}p{1.7cm}}
\hline
Model &Encoder-only pretraining &Encoder-Decoder pretraining &Encoder structure-awareness &Decoder structure-awareness  \\
\hline
CodeBERT\citep{codebert} &MLM, RTD &- &- &- \\
GraphCodeBERT\citep{guo2020graphcodebert} &MLM, EP, NA &- &DFG &- \\
Transcoder\citep{transcoder} &MLM &DAE, BT &-&- \\
PLBART\citep{plbart} &- & DAE &- & -\\
DOBF\citep{dobf} &- &DOBF &- &- \\
CodeT5\citep{wang2021codet5} &IT &MSP, MIP, NL-PL dual generation &Identifiers &Identifiers  \\
MuST\citep{zhu2022multilingual} &- &DAE, MuST &- &- \\
StructCoder (ours) & &structure-based DAE, NL-PL dual generation &AST, DFG &AST, DFG \\
\hline
\end{tabular}
\label{tab:related}
\end{table*}

\subsection{Pretrained Transformers for Code} 
The recent state-of-the-art results on most natural language generation tasks are obtained by pretraining huge deep learning models on large datasets with carefully designed pretext tasks. 
Since code generation is very similar to text generation and there is abundant unsupervised code data available through open source code repositories, pretraining code generation models using similar pretext tasks has been successful. Most recent state-of-the-art pretrained models for code utilize the Transformer \citep{vaswani2017attention} architecture and are discussed below.

CodeBERT \citep{codebert} performs encoder-only pretraining using Masked Language Modeling and Replaced Token Detection as pretext tasks on the CodeSearchNet dataset. 
Transcoder \citep{transcoder} is an unsupervised translation model which pretrains both encoder and decoder using Denoising Autoencoding and Back-Translation with only monolingual datasets.
PLBART \citep{plbart} is pretrained with DAE objective using 680M Java and Python functions. 
DOBF \citep{dobf} attempts to understand code structure with a deobfuscation pretext task where every occurrence of a sampled identifier is replaced by an uninformative token. 
Code Transformer \citep{zugner2021language} modifies the attention computations in the encoder according to AST-based distances.
CodeT5 \citep{wang2021codet5} pretrains a T5 model \citep{raffel2020exploring} with code data in 8 programming languages. In contrast to PLBART, which treats code data as plain sequences, CodeT5 includes identifier-aware objectives in the training, which helps maintain the correctness of the code. However, CodeT5 does not include any structural information of the code in training. 
\citet{zhu2022multilingual} improve code translation performance by introducing a fine-grained snippet-level translation task during pretraining. 
GraphCodeBERT \citep{guo2020graphcodebert} utilizes code structure in the form of Data Flow Graph (DFG) which contains semantic information as opposed to the syntatic information in AST. However, the decoder is completely unaware of the code structure in all of the above methods. 
\textit{Our model advances the domain of code generation by being the first one to train a structure-aware Transformer encoder and decoder by modeling both syntax and data flow.} 
A summary of the pretext tasks and code structures used by the above Transformer-based methods along with our approach is provided in Table \ref{tab:related}. 

\section{StructCoder} \label{sec:proposed}
StructCoder is a Transformer based encoder-decoder model where both encoder and decoder are structure-aware. We build our model using T5 architecture and add the relevant components for modeling code structure. For code inputs,
the encoder (refer to Section \ref{sec:encoder}) inputs the tokenized source code sequence along with its AST and DFG and employs structure-aware self-attention. The structure-aware decoder (refer to Section \ref{sec:decoder}) simultaneously learns to generate the target code sequence as well as to perform target AST and DFG related tasks. The notations used to describe our methodology in this section are summarized in Table \ref{tab:symbols}.

\begin{table*}[!h]
    \caption{Notations used in this paper.}
    \label{tab:symbols}
    \centering
\footnotesize
\begin{tabular}{cp{8cm}}
\hline
Notation & Definition \\
\hline
$S=(s_1,...,s_{|S|})$ & Input/source token sequence\\
$T=(t_1,...,t_{|T|})$ & Output/target token sequence\\
$\mathcal{T}=(N,N_{leaf},r,p(.),L^{ast})$ & AST (Abstract Syntax Tree) with root node $r$ \\
$N$ & Set of AST nodes \\
$N_{leaf}=\{l_1,...,l_{|N_{leaf}|}\}$ & Set of AST leaf nodes \\
$|l_i|$ & No. of nodes on the root-$l_i$ path in AST\\
$p:N\textbackslash \{r\}\longrightarrow N$ & Parent node mapping in AST\\
$L^{ast}\in \{0,1\}^{(\text{\# code tokens})\times|N_{leaf}|}$ & Token-leaf linking matrix. \\
$\mathcal{Y}$ & Set of node types \\
$n.type \in \mathcal{Y}$ & Type of a node $n$ in AST \\
$E_{type}(.)$ & Node type embedding \\
$E_{height}(.)$ & Node height embedding \\
$\mathcal{G}=(V,D,L^{dfg})$ & DFG (Data Flow Graph) \\
 $V=\{v_1,v_2,...,v_{|V|}\}$ & Set of variables in DFG\\
$D\in \{0,1\}^{|V|\times |V|}$ & DFG adjacency matrix\\
$L^{dfg}\in \{0,1\}^{(\text{\# code tokens})\times |V|}$ & Token-variable linking matrix\\
$E_x\in \mathbb{R}^d$ & Embedding of x \\
$A(.,.)$ & Attention score before softmax \\
$W_q,\,W_k$ & Query and Key projection matrices\\
$\phi:\mathbb{Z}_{\geq 0}\longrightarrow \mathbb{R}$ & Relative position embedding\\
$h_i\in \mathbb{R}^d$ & Hidden state from the last decoder layer at $i^{th}$ position\\
$l_{t_i}$ & Leaf node containing code token $t_i$\\
$p_i\in[0,1]^{|\mathcal{V}|}$ & Predicted probability of each token in vocabulary $\mathcal{V}$ at $i^{th}$ position \\
$p_{ik}^{ast}\in[0,1]^{|\mathcal{Y}|}$ & Predicted probability of each node type for the $k^{th}$ AST node on the root-$l_{t_i}$ path.  \\
$p^{dfg}_{ij}\in[0,1]$ & Predicted probability of data flow from $t_j$ to $t_i$. \\
$y_{ij}\in\{0,1\}$ & Ground truth data flow indicator from $t_j$ to $t_i$ \\
$\mathcal{L}_{lm}$ & Language modeling loss\\
$\mathcal{L}_{app}$ & AST Paths Prediction (APP) loss\\
$\mathcal{L}_{dfp}$ & Data Flow Prediction (DFP) loss\\
$\alpha_1,\,\alpha_2$ & DFP loss weight,  APP loss weight \\
\hline
\end{tabular}
\end{table*}

\subsection{Preliminaries}
A \textbf{Code} can be a function or a program, and is represented as a sequence of tokens $S=(s_1,...,s_{|S|})$. 
A code $S$ has a corresponding \textbf{AST} represented as $\mathcal{T}=(N,N_{leaf},r,p(.),L^{ast})$, where $N$ is the set of nodes in the AST, $N_{leaf}=\{l_1,...,l_{|N_{leaf}|}\}\subset N$ is the subset of leaf nodes, $r\in N$ is the root node, $p:N\textbackslash \{r\}\longrightarrow N$ is a mapping such that $p(n)$ denotes the parent of node $n$, and $L^{ast}\in \{0,1\}^{|S|\times|N_{leaf}|}$ is a linking matrix such that $L^{ast}_{ij}=1$ if and only if token $s_i$ is part of leaf $l_j$. Each node $n\in N$ has a type denoted by $n.type$. We use the tree-sitter library\footnote{https://github.com/tree-sitter/py-tree-sitter} to parse codes and generate syntax trees according to a context-free grammar for each programming language.

A code $S$ also has a corresponding
\textbf{DFG} represented as $\mathcal{G}=(V,D,L^{dfg})$, where $V=\{v_1,v_2,...,v_{|V|}\}$ is the set of variables extracted from code $S$, and $D\in \{0,1\}^{|V|\times |V|}$ is the adjacency matrix where $D_{ij}=1$ if and only if value of $v_i$ is directly obtained from $v_j$, and $L^{dfg}\in \{0,1\}^{|S|\times |V|}$ is a linking matrix such that $L^{dfg}_{ij}=1$ if and only if variable $v_j$ is derived from token $s_i$. For extracting DFG, we utilize the implementation\footnote{https://github.com/microsoft/CodeXGLUE/blob/main/Code-Code/code-to-code-trans/evaluator/CodeBLEU/parser/DFG.py} of \citet{ren2020codebleu} where the tree-sitter generated AST is traversed to recursively identify the variables and data flow relations between them using a language-specific deterministic function.

The goal of code translation is to transform a code $S=(s_1,...,s_{|S|})$ in a source language to code $T=(t_1,...,t_{|T|})$ in a different target language such that the translated code $T$ is semantically equivalent to the input code $S$. In text-to-code generation, the goal is to generate target code $T$ from a natural language description.

\begin{figure*}[!ht]
    \centering
    \includegraphics[scale=0.58, trim=1cm 0.8cm 0 0.2cm]{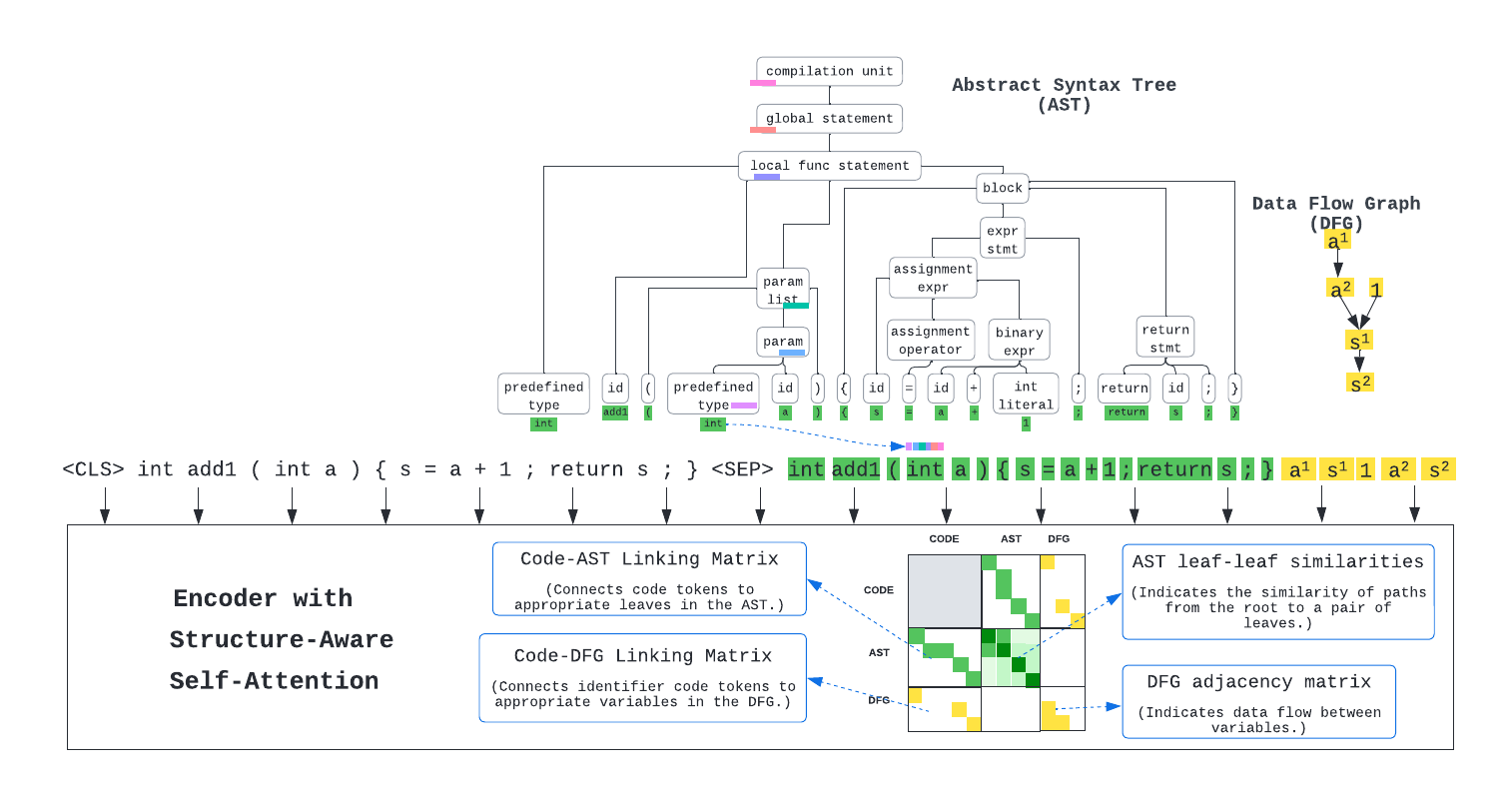}
    \caption{Structure-aware encoder: The input sequence to the encoder consists of source code concatenated with the AST leaves and DFG variables, where the AST leaves are embedded using the root-leaf paths in the AST. The modified structure-aware self-attention mechanism of this Transformer encoder utilizes code-AST/DFG linking information, leaf-leaf similarities in the AST, and the (asymmetric) DFG adjacency matrix to compute the attention matrix.}
    \label{fig:enc}
\end{figure*}

\subsection{Structure-Aware Encoder} \label{sec:encoder}
Given source code $S$, its corresponding AST $\mathcal{T}$, and DFG $\mathcal{G}$, 
the input sequence to the encoder is  $$\langle CLS \rangle,s_1,..,s_{|S|},\langle SEP \rangle,l_1,...,l_{|N_{leaf}|},v_1,...,v_{|V|}$$
which consists of the code tokens, special tokens $\langle CLS\rangle$ and $\langle SEP\rangle$, AST leaves, and DFG variables. For text input, the leaves and variables are simply ignored in the input. The encoder architecture is illustrated in Fig. \ref{fig:enc} and is described in detail below.

\subsubsection{Input Embedding} As StructCoder consists of a Transformer encoder, each token in the input sequence has to be embedded in $\mathbb{R}^d$.
We embed the code tokens along with special tokens by using a lookup table, and use a unique embedding for all DFG variables. The DFG information will be used by the encoder in structure-aware self-attention. We compute the embedding of a leaf $l$ in an AST as a function of the path from the root to the leaf $l$ in the AST.

Let $(r_1,r_2,...,r_{|l|})$ be the nodes on the path from root $r=r_1$ to leaf $l=r_{|l|}$. We utilize node-type embedding $E_{type}(\cdot)\in \mathbb{R}^d$ to encode a node's syntax along with a node-height embedding $E_{height}(\cdot)\in \mathbb{R}^d$ to encode the order of nodes on this path. 
The leaf embedding $E(l)$ is computed as
\begin{align}
    E(l) = \sum_{i=1}^{|l|} E_{type}(r_i.type) \odot E_{height}(|l|-i) \;\; \in \mathbb{R}^d
\end{align}
where $\odot$ denotes element-wise multiplication.


\subsubsection{Structure-aware Self-attention} Since the input contains DFG and AST which consist of structural information, the traditional attention computation using (relative) positional embeddings which capture sequential ordering information is not sufficient. Hence, we propose structure-aware self-attention which computes attention scores between tokens based on the structural relations between them.

\noindent
\underline{Code-code:} Following T5, we compute attention scores (before softmax) between code tokens by adding the query-key dot product with weights $W_q,W_k\in \mathbb{R}^{d_k\times d}$ and a lookup embedding $\phi:\mathbb{Z}_{\geq 0}\longrightarrow \mathbb{R}$ for relative position. Denoting embedding of $x$ by $E_x$, we have
\begin{align}
    A(s_i,s_j) &= E_{s_i}^TW_q^TW_kE_{s_j} + \phi({|i-j|})  \label{eq:code-code}
\end{align}

\noindent
\underline{Leaf-leaf:} To calculate attention scores between leaves, we introduce a similarity-based transformation to replace the relative positional embedding in equation \ref{eq:code-code}. Let $(r^{i}_1,...,r^{i}_{|l_i|})$ be the nodes on the path from root to leaf $l_i$.
We define similarity between two leaves $l_i$ and $l_j$ as 
\begin{align}
    sim(l_i,l_j) &= log \left (1 + \frac{\bigg( \sum \limits_{k=1}^{min(|l_i|,|l_j|)}\mathbbm{1}(r^{i}_k=r^{j}_k)\bigg )^2}{|l_i||l_j|}  \right ) 
    \label{eq:sim}
\end{align}
which is based on the number of common nodes on the paths from root to leaves $l_1$ and $l_2$. The $log$ transformation is used to reduce the skewness of the distribution of similarity values. The attention scores between leaves are then computed as follows.
\begin{align}
    A(l_i,l_j) &= E_{l_i}^TW_q^TW_kE_{l_j} + (w_a \; sim(l_i,l_j)+w_b)
\end{align}
where $w_a,w_b\in \mathbb{R}$ are learnable parameters.

\vspace{0.04in}
\noindent
\underline{Variable-variable:} Following \citet{guo2020graphcodebert}, the attention scores between DFG nodes are computed using only the query-key dot product. They are set to $-\infty$ if corresponding edges are absent in the DFG. 
\begin{align}
    A(v_i,v_j) &= \begin{cases} E_{v_i}^TW_q^TW_kE_{v_j} & \text{if} \; D_{ij}=1 \\ -\infty & else
    \end{cases} 
\end{align}

\noindent
\underline{Code-leaf/variable:} For interaction between code tokens and AST leaves (or DFG variables), we only compute the query-key dot product and do not use any positional information. Inspired by the work of \citet{guo2020graphcodebert}, we set the attention score to $-\infty$ for cases where the leaf (or variable) is not linked to the code token. We show the equations only for interactions between code tokens and leaves as those for interactions between code tokens and variables are similar.
\begin{align}
    A(s_i, l_j) &= \begin{cases} E_{s_i}^TW_q^TW_kE_{l_j} & \text{if} \; L^{ast}_{ij}=1 \\ -\infty & else
    \end{cases}; \quad A(l_j, s_i) &= \begin{cases} E_{l_j}^TW_q^TW_kE_{s_i} & \text{if} \; L^{ast}_{ij}=1 \\ -\infty & else
    \end{cases} \label{eq:6}
\end{align}

\subsection{Structure-Aware Decoder} \label{sec:decoder}
\begin{figure}[!h]
    \centering
    \includegraphics[scale=0.65, trim=0.6cm 0.7cm 0 0]{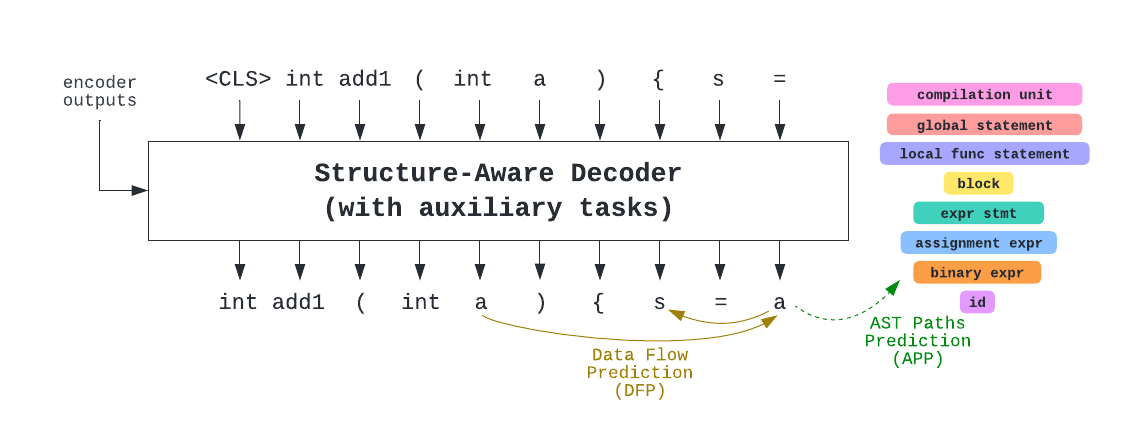}
    \caption{Structure-aware decoder generates the next token in the target code as well as predicts the node types on the root-leaf path to the leaf containing this token in the target AST and also the DFG edges incident on this token.}
    \label{fig:dec}
\end{figure}

The decoder in StructCoder constitutes the original T5 decoder with additional layers at the end for AST paths prediction and data flow prediction tasks that are introduced in this section.
Fig. \ref{fig:dec} illustrates the structure-aware decoder which predicts the next target code token along with the AST root-leaf path to this token and the data flow relations between this token and all past tokens. The addition of these auxiliary tasks does not increase the number of generated tokens, which is important since the decoding is done in an autoregressive manner.

Let $h_1,h_2,...,h_{|T|}$ be the hidden states generated by the Transformer decoder. Decoders of existing transformer models including T5 employ 
a linear layer with weights $W\in \mathbb{R}^{ |\mathcal{V}| \times d}$ followed by softmax transformation to extract a probability distribution $p_i$ on the token vocabulary space $\mathcal{V}$ for the $i^{th}$ position.
\begin{align}
p_i = softmax\,(Wh_{i})    
\end{align}
And the sequence generation task is trained using language modeling loss as shown below for one sample.
\begin{align}
    \mathcal{L}_{lm} = -\frac{1}{|T|}\sum_{i=1}^{|T|}log\;p_i(t_i)
\end{align}
where  $p_i(t_i)$ refers to the predicted probability for true target token $t_i$ at the $i^{th}$ position.
	
In addition to sequence generation, StructCoder also learns target syntax using AST paths prediction task, and learns to match target DFG using a data flow prediction task.

\subsubsection{AST Paths Prediction (APP)} In this task, the goal is to encourage the decoder to be aware of all root-leaf paths in the target AST. Since the type attribute of a node captures important syntactic information, we predict the type of each ancestor on each root-leaf path.

Let $l_{t_i}$ be the leaf node containing the $i^{th}$ target token $t_i$ and let $(r^{i}_1,...,r^{i}_{|l_{t_i}|})$ be the nodes on the root-$l_{t_i}$ path.
To predict type of node $r^i_k$ (which is at height $|l_{t_i}|-k$ in the tree), we use a linear layer with weights $W_{ast\,(|l_{t_i}|-k)}\in \mathbb{R}^{|\mathcal{Y}|\times d}$ followed by a softmax transformation to predict a probability distribution on the set of node types $\mathcal{Y}$.
\begin{align}
    p_{ik}^{ast} = softmax( W_{ast\,(|l_{t_i}|-k)}h_i )
\end{align}
The APP cross-entropy loss for a sample is given by
\begin{align}
    \mathcal{L}_{app} = -\frac{1}{|T|(\sum_i|l_{t_i}|)}\sum_{i=1}^{|T|} \sum_{k=1}^{|l_{t_i}|} log\; p^{ast}_{ik}(r_k^{i}.type)
\end{align}


\subsubsection{Data Flow Prediction (DFP)} In this task, the decoder learns to predict all the data flow edges in target code.
The probability $p^{dfg}_{ij}$ of data flow from $j^{th}$ to $i^{th}$ position in target code sequence is computed using an asymmetric transformation (since data flow is directed) as
\begin{align}
    p^{dfg}_{ij} = \sigma\, (h_i^TU_{dfg}^TV_{dfg}h_j+u_{dfg}^Th_i+v_{dfg}^Th_j+w_{dfg}) 
\end{align}
where $\sigma(.)$ denotes the sigmoid function.
Suppose $\mathcal{G}=(V,D,L)$ is the true target DFG.
There is a data flow from $j^{th}$ to $i^{th}$ position in target sequence if and only if 
``target DFG contains variables $v_{j'},\,v_{i'}$ such that variable $v_{j'}$ is derived from $t_j$, variable $v_{i'}$ is derived from $t_i$, and value of variable $v_{i'}$ is derived from $v_{j'}$".
Thus, the DFP loss for a sample can be written as
\begin{align}
    \mathcal{L}_{dfp} &= -\frac{1}{|T|^2} \sum_{i=1}^{|T|} \sum_{j=1}^{|T|} \big\{ \,y_{ij}\, log\,p^{dfg}_{ij} 
    + (1-y_{ij})\, log\,(1-p^{dfg}_{ij}) \,\big\} \nonumber \\
    &\text{where}\quad 
    y_{ij} = \mathbbm{1}(\exists \,v_{i'},\,v_{j'}\in V \,\,\text{such that} \,\,  D_{i'j'}=  L^{dfg}_{ii'}=L^{dfg}_{jj'}=1)
\end{align}

The overall loss function for training StructCoder (given below) is a combination of the language modeling objective, and the APP and DFP losses with weights $\alpha_1$ and $\alpha_2$, i.e. $\mathcal{L} = \mathcal{L}_{lm} + \alpha_1\mathcal{L}_{app} + \alpha_2\mathcal{L}_{dfp}$.

\subsection{Pretraining}
We pretrain StructCoder on the CodeSearchNet \citep{husain2019codesearchnet} dataset\footnote{\url{https://github.com/github/CodeSearchNet}} containing about 2M code and comment pairs, with a structure-based DAE task along with NL-PL bimodal dual generation to generate code from text and vice-versa. 
For the denoising task, we corrupt random spans in the code sequence by replacing them with $\langle MASK\rangle$ or a random token or deleting them. The span lengths are sampled from a Poisson distribution with a mean of 12 tokens. We corrupt 35\% of the code tokens in total, similar to \cite{plbart}. To improve the understanding of code structure, we also randomly drop 35\% of the DFG variables and AST leaves, and 35\% of the ancestors for each leaf from the input to StructCoder. The model is then trained to predict the uncorrupted code along with the AST root-leaf paths and data flow edges. We initialize our model for pertaining with CodeT5's weights (for faster pretraining) except for the AST and DFG related weights, which are randomly initialized.

\subsection{Implementation Details}
We implement StructCoder by extending the CodeT5-base architecture containing 12 T5 blocks with hidden dimension 768, and 12 attention heads in each block. StructCoder comprises a total of 224M trainable parameters, while CodeT5-base contains 223M.  We employ the  AdamW \cite{loshchilov2017decoupled} optimizer with a learning rate of 2e-4 for pretraining and 1e-5 for finetuning. We ran the pretraining for 175K batches with a batch size of 20 code-comment pairs. For finetuning, we used batch sizes of 25, 32, and 20 for CodeXGLUE translation, CONCODE, and APPS datasets, respectively. The fine-tuning was run for 50K, 300K, and 40K batches for the three datasets, respectively. The loss weights of auxiliary tasks, $\alpha_1$ and $\alpha_2$, are both set to 0.1.
To facilitate minibatch training with available resources, we set the maximum number of DFG variables in the input to 65, the maximum number of AST leaves to 250, and the maximum root-leaf path length to 17 (by trimming paths from the root’s side). We set the maximum source length (no. of code/text tokens) to 400 for pretraining, 320 for translation, 320 and 600 for text-to-code generation
on CONCODE and APPS. We set the maximum target length to 400 for pretraining, 256 for translation, 150 and 512 for
text-to-code generation on CONCODE and APPS, respectively. We implement our model using the PyTorch \cite{paszke2019pytorch} and Hugging Face \cite{wolf-etal-2020-transformers} libraries. Additional implementation and experimental setup details are provided in the Appendix.

\section{Experiments} \label{sec:experiments}
We evaluate StructCoder on the code translation and text-to-code generation tasks from the CodeXGLUE \footnote{\url{https://github.com/microsoft/CodeXGLUE}} \citep{codexglue} benchmark, and on the text-to-code generation task from the APPS benchmark \citep{hendrycks2021measuring}, and compare with previously published results on these tasks.\footnote{We did not include CODEGEN \cite{nijkamp2022codegen} and Incoder \cite{fried2022incoder} in the baselines because these models are trained on much bigger datasets and/or use much larger architectures and it is unfair to compare them with our model and the other baselines used in this paper.}
For CodeXGLUE tasks, we use the metrics from the CodeXGLUE leaderboard which include (i) BLEU \citep{papineni2002bleu} score which measures n-gram overlap, (ii) exact match (xMatch) which checks if the prediction is the same as ground truth, and (iii) CodeBLEU \citep{ren2020codebleu} which combines BLEU score with keywords-based weighted n-gram match as well as syntax and semantic matches based on AST and DFG. APPS evaluates generated codes based on test cases where the evaluation metrics include (i) `test case average' which is the average percentage of test cases passed, and (ii) `strict accuracy' which is the percentage of generated codes that pass all test cases.

\subsection{Code Translation}
The code translation dataset from CodeXGLUE consists of two tasks for translating between Java and C\# functions in either direction and contains 10K training samples, 500 validation samples, and 1000 test samples. Table \ref{tab:code_trans} presents the results of StructCoder alongside the baselines on the two code translation tasks. 
The Naive Copy baseline simply copies source code to target, and the Transformer model does not include any pretraining.  RoBERTa (code) \citep{codexglue}, CodeBERT, and GraphCodeBERT involve encoder-only pretraining while PLBART and CodeT5 incorportate encoder-decoder pretraining like StructCoder.
StructCoder achieves the best results on the two translation tasks which can be attributed to the structure-aware encoder-decoder design of our model.
From Table \ref{tab:code_trans}, we observe that the encoder-decoder pretraining of PLBART, CodeT5, and StructCoder is very beneficial to code translation. Also, the encoder-only pretrained models improve over Transformer by a huge margin. 
GraphCodeBERT which utilizes DFG offers minor improvements over CodeBERT and we also observed in our ablation study that DFG-related components contribute less to the performance gains of StructCoder compared to AST-related components.

\begin{table*}[!h]
    \centering
    \caption{Results on code translation tasks from CodeXGLUE benchmark. (*Since CodeT5 is a competitive baseline and did not report CodeBLEU in their paper, we tested this model using their finetuned checkpoint and provided the results.)
    }
\begin{tabular}{lcccccc}
\hline
& \multicolumn{3}{c}{Java-C\#}                     & \multicolumn{3}{c}{C\#-Java}                     \\
\cmidrule(l{4pt}r{4pt}){2-4} \cmidrule(l{4pt}r{4pt}){5-7}
               & BLEU           & xMatch         & CodeBLEU       & BLEU           & xMatch         & CodeBLEU       \\
               \hline
Naive Copy     & 18.54          & 0.00           & 42.20              & 18.69          & 0.00           & 34.94              \\
Transformer    & 55.84          & 33.00          & 63.74          & 50.47          & 37.90          & 61.59          \\
RoBERTa (code) & 77.46          & 56.10          & 83.07          & 71.99          & 57.90          & 80.18          \\
CodeBERT       & 79.92          & 59.00          & 85.10          & 72.14          & 58.80          & 79.41          \\
GraphCodeBERT  & 80.58          & 59.40          & -              & 72.64          & 58.80          & -              \\
PLBART         & 83.02          & 64.60          & 87.92          & 78.35          & 65.00          & 85.27          \\
CodeT5*         & 83.88          & 64.70          & 87.38          & 79.71          & 67.50          & 85.51          \\
StructCoder    & \textbf{84.43} & \textbf{66.90} & \textbf{88.19} & \textbf{80.43} & \textbf{68.70} & \textbf{85.98} \\
\hline
\end{tabular}
    \label{tab:code_trans}
\end{table*}

\subsection{Text-to-Code Generation}
The text-to-code generation task from CodeXGLUE uses the CONCODE \citep{iyer2018mapping} dataset and the goal here is to generate a Java function given a natural language description. This dataset contains 100K training samples, 2K validation samples, and 2K test samples. Table \ref{tab:code_gen} presents the results of our model alongside the baselines on the text-to-code generation task.
Among the baselines, GPT-2 \citep{radford2019language} is pretrained on natural language to predict next token, CodeGPT \citep{codexglue} is pretrained from scratch like GPT-2 but using code data, CodeGPT-adapted \citep{codexglue} is pretrained from GPT-2 initialization using code data, and CoTexT \citep{phan2021cotext} pretrains the T5 model further on code data using MSP objective.
The decoder-only baselines which include GPT-2 based models are outperformed by the rest which are all encoder-decoder models.
StructCoder again achieves the best performance on all metrics for the text-to-code generation task.

\begin{table}[!h]
    \centering
    \caption{Results on text-to-code generation task from CodeXGLUE benchmark.}
    \begin{tabular}{lccc}
         \hline
    &BLEU &xMatch &CodeBLEU \\
    \hline
    GPT-2       & 25.37     & 17.35                                         & 29.69                        \\
CodeGPT         & 28.69                    & 18.25                      & 32.71                        \\
CodeGPT-adapted & 32.79                    & 20.10                      & 35.98                        \\
PLBART          & 36.69                    & 18.75                      & 38.52                        \\
CoTexT          & 37.40                    & 20.10                      & 40.14                        \\
CodeT5          & 40.73                    & {22.30}                     & 43.20                        \\
StructCoder & \textbf{41.35} & \textbf{22.40} & \textbf{44.44} \\
    \hline
    \end{tabular}
    \label{tab:code_gen}
\end{table}

\begin{table*}[!h]
    \caption{Results on the APPS dataset along with model size in \#billion parameters. The results for GPT-2 models were obtained from \cite{hendrycks2021measuring}.}
    \label{tab:apps}
    \centering
\begin{tabular}{p{1.5cm}ccccccc}
\hline
& & \multicolumn{3}{c}{Test case average}  & \multicolumn{3}{c}{Strict accuracy}    \\ \hline
&Model size & Intro & Interview & Competition & Intro & Interview & Competition \\ \hline
GPT-2 &0.1B                         & 5.64         & 6.93      & 4.37        & 1            & 0.33      & 0           \\
CodeT5 &0.2B &9.50 &6.03 &2.51 &1.70 &0.37 &0     \\
StructCoder &0.2B                   & \textbf{10.22}        & 7.50         &  3.18     & \textbf{2.5}         & \textbf{0.70}         &  \textbf{0.2}         \\
GPT-2 &1.5B & 7.40          & \textbf{9.11}      & \textbf{5.05}        & 1.3          & \textbf{0.70}       & 0           \\
\hline
\end{tabular}
\end{table*}

APPS \citep{hendrycks2021measuring} is a text-to-code generation benchmark in python which evaluates generated codes based on test cases. The inputs here contain detailed questions and possibly some starter code as well.
The dataset contains 10K problems equally divided into train and test splits.  The test set contains 1K introductory level, 3K interview level, and 1k competition level problems. Table \ref{tab:apps} shows the results of StructCoder, CodeT5, and GPT-2 \citep{hendrycks2021measuring} models of two sizes. 
These GPT-2 models were pretrained exclusively on python code from GitHub which gives them an edge in this particular task.  The `strict accuracy' metric is more important than the `test case average' as it does not give partial credit to a generated code that does not pass all test cases.
StructCoder achieves the best `strict accuracy' on all subsets, notably outperforming the bigger GPT-2 model which is about 7 times the size of StructCoder.


\subsection{Model Analysis} \label{sec:limitations}
\subsubsection{Ablation Study} To emphasize the importance of the novel structure-based components introduced in this work, we conducted an ablation study on the two code translation tasks from CodeXGLUE. 
For this experiment, we used a smaller T5 architecture with hidden dimension 256, 5 encoder and decoder layers, and 8 heads in each multi-head attention layer.
The ablated models we tested here include the smaller T5 model
(i) without any of the proposed structure-based components (No structure (baseline)); (ii) enabling DFG in the encoder (DFG(enc)); (iii) enabling Data Flow Prediction task in the decoder (DFG(dec)); (iv) enabling AST in encoder (AST(enc)); (v) enabling AST Paths Prediction task in the decoder (AST (dec)); (vi) enabling all proposed structure-based components/tasks; and (vii) adding structure-based DAE pretraining to (vi). We report the CodeBLEU metric along with its different components for each of these models in Table \ref{tab:ablation_java2cs}. Among the different components of the CodeBLEU metric, weighted BLEU gives more weight to programming language keywords, AST match computes the percentage of subtrees in the ground truth target AST that occur in the generated code, and DFG match computes the percentage of DFG edges in the ground truth that occur in the generated code.

Enabling each of the four [(ii)-(v)] structure-based components individually results in an increase in AST match and data flow match metrics over the baseline [(i)] in most of the cases. The DFG components in the model [(ii),(iii)], however, do not seem to always increase BLEU and weighted BLEU scores. Among the four components [(ii)-(v)], enabling AST paths prediction task [(v)] yields the best BLEU and weighted BLEU, and modeling AST in the input [(iv)] yields the best AST match. Enabling all the components [(vi)] gives the best results on AST match, data flow match, and overall CodeBLEU scores. We also observed that structure-based DAE pretraining [(vii)] led to significant performance gains on both tasks.


\begin{table*}[!ht]
    \centering
    \caption{CodeBLEU and its different components on Java-C\# and C\#-Java translation for the validation sets by adding the proposed structure-based components to a smaller T5 model. (`enc' and `dec' indicate whether the proposed structure-based components/tasks were included in the encoder and decoder, respectively. AST stands for Abstract Syntax tree, DF for Data Flow, and `wBLEU' for weighted BLEU.)
    }
\footnotesize
\begin{tabular}{p{3.9cm}p{0.6cm}p{0.6cm}p{0.6cm}p{0.6cm}p{0.6cm}p{0.6cm}p{0.6cm}p{0.6cm}p{0.6cm}p{0.6cm}}
\hline
 & \multicolumn{2}{c}{BLEU}     & \multicolumn{2}{c}{wBLEU} & \multicolumn{2}{c}{AST match} & \multicolumn{2}{c}{DF match} & \multicolumn{2}{c}{CodeBLEU} \\
\hline
& J-C & C-J & J-C & C-J & J-C & C-J & J-C & C-J & J-C & C-J \\
\hline
(i) No structure (baseline) & 60.00 & 54.46 & 61.85 & 55.64 & 78.10 & 74.79 & 72.41 & 64.92 & 68.09 & 62.45\\
(ii) DFG (enc) & 59.20 & 54.38 & 61.66 & 55.60 & 78.20 & 75.56 & 73.20 & 66.89 & 68.07 & 63.11\\
(iii) DFG (dec) & 61.25 & 54.45 & 62.78 & 55.58 & 78.72 & 76.11 & 73.08 & 66.39 & 68.96 & 63.13\\
(iv) AST (enc) & 60.78 & 54.70 & 62.21 & 55.87 & 79.15 & 76.67 & 73.69 & 67.02 & 68.96 & 63.57\\
(v) AST (dec) & \underline{61.76} & \underline{56.40} & \underline{63.16} & \underline{57.42} & 78.72 & 75.65 & 73.91 & 64.81 & 69.39 & 63.57\\
(vi) DFG (enc, dec), AST (enc, dec) & 61.51 & 55.43 & 62.89 & 56.43 & \underline{79.71} & \underline{77.36} & \underline{74.12} & \underline{67.80} & \underline{69.56} & \underline{64.26}\\
(vii) DFG (enc, dec), AST (enc, dec), \& structure-based DAE pretraining & \textbf{80.58} & \textbf{77.09} & \textbf{81.17} & \textbf{77.62}	&\textbf{89.03} & \textbf{89.28 }&\textbf{86.87}	& \textbf{87.24}& \textbf{84.50}  & \textbf{82.67}
\\
\hline
\end{tabular}
    \label{tab:ablation_java2cs}
\end{table*}

\subsubsection{Auxiliary Tasks} We measure the performance of StructCoder on the auxiliary tasks of APP (AST Paths Prediction) and DFP (Data Flow Prediction) as follows. When predicting the next target token, we use the ground truth for target sequence until the previous step as input to the decoder. The decoder then predicts the next token as well as the DFG edges incident on this token and the types of nodes on the path from root to the leaf node containing this token in the AST. 
On Java-C\# translation, StructCoder achieves 94\% accuracy on APP task and 94.7\% average precision on DFP task where positive class prevalence is just 0.8\%.
On C\#-Java translation, StructCoder achieves 96.3\% accuracy on APP task and 82.9\% average precision on DFP task where positive class prevalence is just 0.5\%. For both the translation tasks, there are 298 classes for node type in APP task.

\begin{figure*}[!h]
    \centering
    \includegraphics[scale=1, trim=0cm 0.8cm 0 0.2cm]{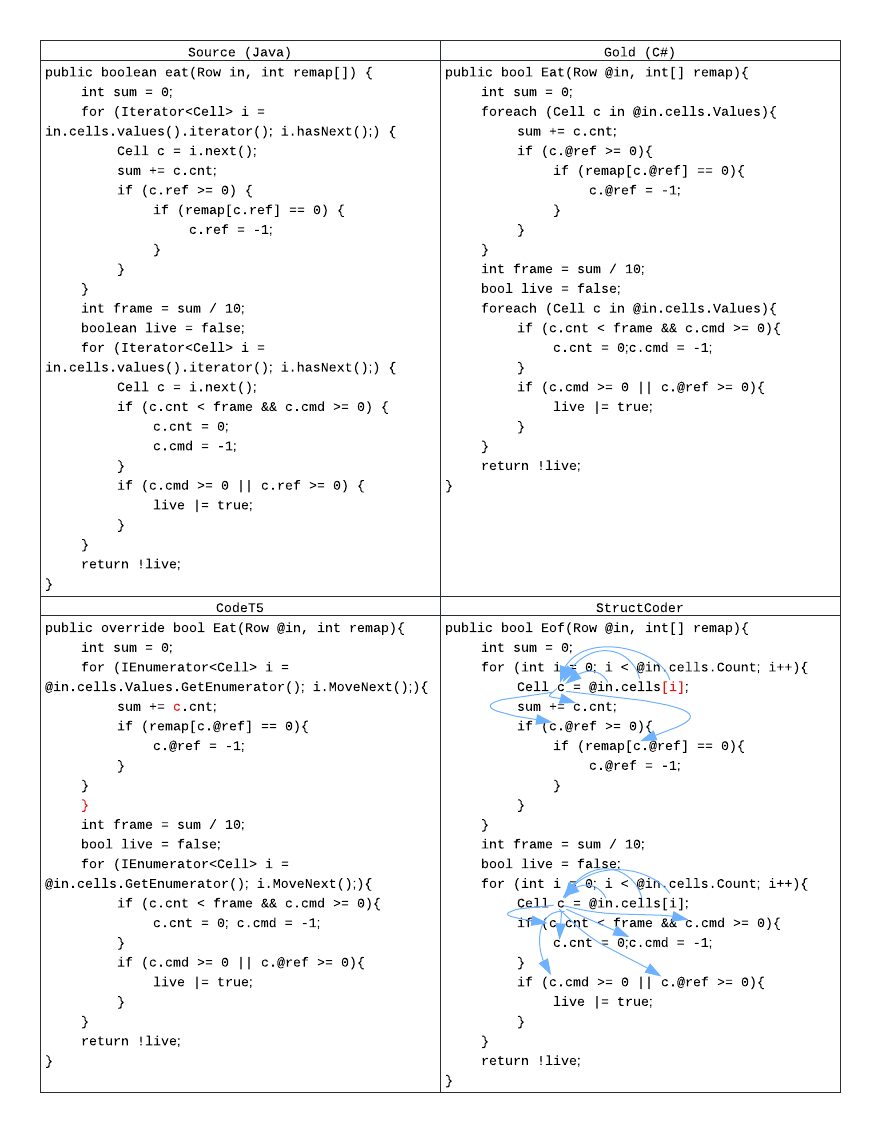}
    \caption{Case study: An example from Java-C\# translation task comparing the outputs from StructCoder and CodeT5. StructCoder only makes one error by assuming that `cells' is an array of `Cell' objects instead of dictionary with values of type `Cell'. CodeT5, however, misses the first `if' statement, produces unbalanced `\}', and does not define variabe `c'. The blue arrows in StructCoder output show the correctly predicted (probability $> 97^{th}$ percentile) data flow edges incident on variable `c'.)}
    \label{fig:case}
\end{figure*}

\subsubsection{Case Study}
Fig. \ref{fig:case} shows an example from Java-C\# translation task with predictions from StructCoder and the best baseline CodeT5. We observe that our structure-aware encoder-decoder architecture is able to generate better target code than CodeT5.
Referring to Fig. \ref{fig:case}, CodeT5 generates both the `for' loops with variable `i', leaving variable `c' undefined.
It also misses the first `if' statement and creates a syntax error from unbalanced braces. CodeT5 also translates the type of argument `remap' as an integer instead of an integer array. On the other hand, StructCoder  generates the `for' loops by defining variable `c' and the model predicts (with a probability greater than the $97^{th}$ percentile) most of the DFG edges incident on the variable `c' inside these `for' loops and also in the first `if' statement. The only error in StructCoder's output is the treatment of `@in.cells' as an array of `Cell' objects instead of a Dictionary with Values of type `Cell'. Such errors motivate the design of better models that align the variables and functions between source and target for code translation.
Also, for token `[]' in args, StructCoder correctly predicts the parent node type `array rank specifier'. More examples are included in the Appendix.

\subsubsection{Inference Time}
To analyze the impact of adding the proposed structure-based components on the overall computational complexity experimentally, we measured the inference times on CodeXGLUE translation tasks by including/excluding the different proposed structure-based components. 
We report the results by running inference on a GPU (NVIDIA Tesla P40 with 12288 MiB memory) for 200 samples from the test set using the maximum batch size that can fit on the GPU with a beam size of 10.
The batch sizes used are 6 when AST is included in encoder, 8 when DFG but not AST is included in encoder, 10 when only the code tokens are fed to the encoder. 
We run decoding till maximum target length is reached so that the model's decoded sequence lengths do not impact the inference times. We did not include preprocessing (tokenization, AST, and DFG construction) time while measuring the inference time because preprocessing took negligible time compared to forward pass.\footnote{For all the 200 samples combined, tokenization took 0.27s (0.30s), and AST and DFG construction took 0.62s (0.88s) for codes in Java (C\#).}

\begin{figure*}[!h]
    \centering
    \begin{subfigure}[t]{0.5\textwidth}
        \centering
        \includegraphics[scale=0.6, trim=1cm 0 0 0]{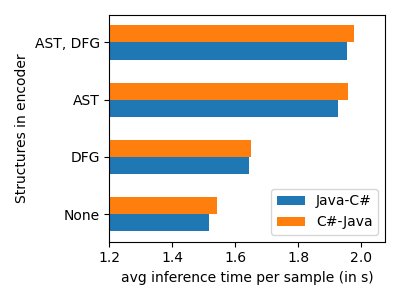}
        \caption{}
        \label{fig:time}
    \end{subfigure}%
    ~ 
    \begin{subfigure}[t]{0.5\textwidth}
        \centering
        \includegraphics[scale=0.6, trim=2cm 0 0 0]{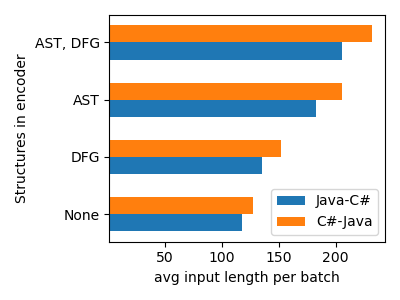}
        \caption{}
        \label{fig:lens}
    \end{subfigure}
    \caption{(a) Inference time (in seconds) per sample averaged over 200 samples, and (b) average input length per batch for the 200 samples in the CodeXGLUE translation tasks for model versions including/excluding AST/DFG related components in the encoder. Since the decoder's structure-based components are not active during inference, we did not consider them in this plot.}
    \label{fig:time_lens}
\end{figure*}

Fig. \ref{fig:time_lens} shows the average inference time per sample and average input length per batch for model versions including and excluding AST and DFG related components in the encoder. Since the decoder's structure-based components are inactive during inference, they do not impact inference time, and hence are not considered here. 
Note that excluding both AST and DFG is equivalent to CodeT5. 
Adding both AST and DFG to encoder increased the inference time by 28\%-29\% compared to using no structures in the encoder, while the input sequence length increased by 75\%-83\%. 
(The increase in inference time being much less than expected may be due to efficient matrix manipulations on GPUs and implementation-specific details of pytorch/huggingface which are out of scope for this work.) 
In our implementation, we compute the full squared attention matrix with diagonal size being equal to the total input length (number of code tokens + number of DFG variables + number of AST leaves), and then mask the attention scores that we want to be zero. But the attention between code tokens and DFG variables / AST leaves, and among the DFG variables is sparse, which motivates more efficient implementations of our method.

\subsection{Performance on Code Summarization}
While the primary focus of this work is on code generation, we have also tested the performance of StructCoder on three languages in the CodeXGLUE summarization benchmark, which is a code-to-text generation task. The results are shown in Table \ref{tab:summa}. StructCoder outperforms CodeT5 by a substantial margin in the case of Go but not in the case of the other languages.

\begin{table}[hbt!]
    \caption{Results on CodeXGLUE summarization}
    \label{tab:summa}
    \centering
\fontsize{9}{11}\selectfont
    \begin{tabular}{lcccc}
    \hline
    & Go & Java & PHP \\
\hline
CodeT5 &19.56 &20.31 &\textbf{26.03} \\
StructCoder &\textbf{24.18} &\textbf{20.39} &25.16 \\
    \hline
    \end{tabular}
\end{table}

\section{Concluding discussion} \label{sec:conclusion}
This work proposes a structure-aware Transformer encoder-decoder model called StructCoder for code generation. Our encoder modifies traditional input embeddings and employs a structure-aware self attention mechanism to model AST and DFG relations in source code, and the decoder is trained to recognize target syntax and data flow using two novel auxiliary tasks to predict the node types on all root-leaf AST paths and data flow edges of target code. We also pretrained our model using a structure-based DAE task to improve its performance. Experiments on code translation and text-to-code generation tasks demonstrate the performance gains of StructCoder over state-of-the-art baselines. We believe that this work would encourage future research in this field to give careful consideration to code structure while building models for code generation.

While automated code generation holds the potential to benefit software development and migration, it comes with inherent risks. 
The models cannot consider constraints like security, efficiency, and modularization when generating code which makes their deployment and maintenance challenging. 
Also, the performance improvements in code generation models largely rely on the scaling-up of both the model and the training, which requires significant computational resources. 
Thus, future research in this area can look into designing more efficient models, and models that generate code conforming to certain preset standards.
\bibliographystyle{ACM-Reference-Format}
\bibliography{6.references}

\newpage
\appendix
\setcounter{table}{0}
\setcounter{figure}{0}
\renewcommand*\thetable{A\arabic{table}}
\renewcommand*\thefigure{B\arabic{figure}}
\section*{\fontsize{12}{15}\selectfont APPENDIX}

\section{More Implementation Details}
We use the CodeT5 tokenizer with a vocabulary size of 32,100. 
As we build upon CodeT5 architecture, both the encoder and decoder of StructCoder contain 12 T5 blocks with hidden dimension 768, and 12 attention heads in each block. During implementation, we only used first 16 bits of the last hidden representation from the decoder to predict DFG links and the next 128 bits for AST paths prediction. 
This is done because the model learns DFP task more easily than APP task and using few bits for these auxiliary tasks prevents overfitting on these tasks. 


\subsection{Finetuning}
For code translation, we ran finetuning with a batch size of 25 for 50K steps. 
For text-to-code generation using CONCODE dataset, we ran finetuning with a batch size of 32 for 300K steps.
To finetune on APPS dataset, we used a batch size of 20 for 40K steps. 
For new AST node types seen during finetuning, we initialized the weights corresponding to these new node types randomly. 
We used beam search with a beam size of 10 for decoding in all finetuning tasks except for the APPS dataset where the beam size was set to 5. 
We ran validation every 500 steps for CodeGLUE translation and every 3000 steps for CONCODE, and  chose the checkpoint with the best CodeBLEU+xMatch score on the validation set for testing. For APPS, which has no validation set, the checkpoint at the end of the training was used for inference. Since CodeT5 does not have published results on the APPS dataset, we finetuned it using the same hyperparameters used by our model.

For the ablation study, the learning rate was set to 2e-4 when training from scratch and 1e-5 for finetuning, and the beam size was set to 5. For the auxiliary tasks of DFP and APP, we use the first 8 and next 32 bits of the last hidden state representation, respectively for the ablation study.

\subsection{Sequence Lengths}
The main paper lists the maximum lengths used for the source and target.
We used the same sequence lengths as StructCoder for finetuning CodeT5 on APPS. 
The results of CodeT5 on CodeXGLUE tasks were borrowed from \citet{wang2021codet5} where the maximum source and target lengths were set to 512 and 256, respectively. 
On the code translation tasks, GraphCodeBERT \citep{guo2020graphcodebert} sets the maximum source and target lengths to 256 and the maximum number of DFG variables to 64.

\subsection{Other Details}
All the hyperparameters discussed above were set either based on CodeT5's implementation, or in rare cases, by observing the progression in validation performance for a few steps, or by choosing the ones with best validation performance after a few trials.
The code for generating ASTs and DFGs is built using tree-sitter \footnote{https://github.com/tree-sitter/py-tree-sitter} and is also adapted from \url{https://github.com/microsoft/CodeBERT/tree/master/GraphCodeBERT}.
The random generators were seeded in the `set\_seed()' function for each experiment.
We ran our experiments on an Ubuntu 18.04 server with 4 RTX 8000 GPUs with 48GB memory on each GPU. We used all 4 GPUs for pretraining and 2 GPUs for the finetuning experiments. 


\subsection{Model Size} Table \ref{tab:sizes} shows the number of parameters in different pretrained models for code. Note that StructCoder is built by adding additional components to CodeT5 for modeling AST and DFG in input and output, with majority of additional parameters coming from the encoder's AST leaves embedding module (381K) and the classification layer of APP (AST Paths Prediction) task (743K).

\begin{table}[hbt!]
    \caption{Number of parameters in various pretrained models}
    \label{tab:sizes}
    \centering
\fontsize{9}{11}\selectfont
    \begin{tabular}{lc}
    \hline
    Pretrained model & \# parameters \\
    \hline
CodeBERT & 125M \\
GraphCodeBERT & 125M \\
CodeGPT-small-java & 126M \\
PLBART & 139M \\
CodeT5 & 223M \\
CoTexT & 223M \\
StructCoder & 224M \\
\hline
    \end{tabular}
\end{table}

\section{Examples}
\begin{figure*}[hbt!]
    \centering
    \includegraphics[scale=0.3, trim=4cm 2cm 0cm 1cm]{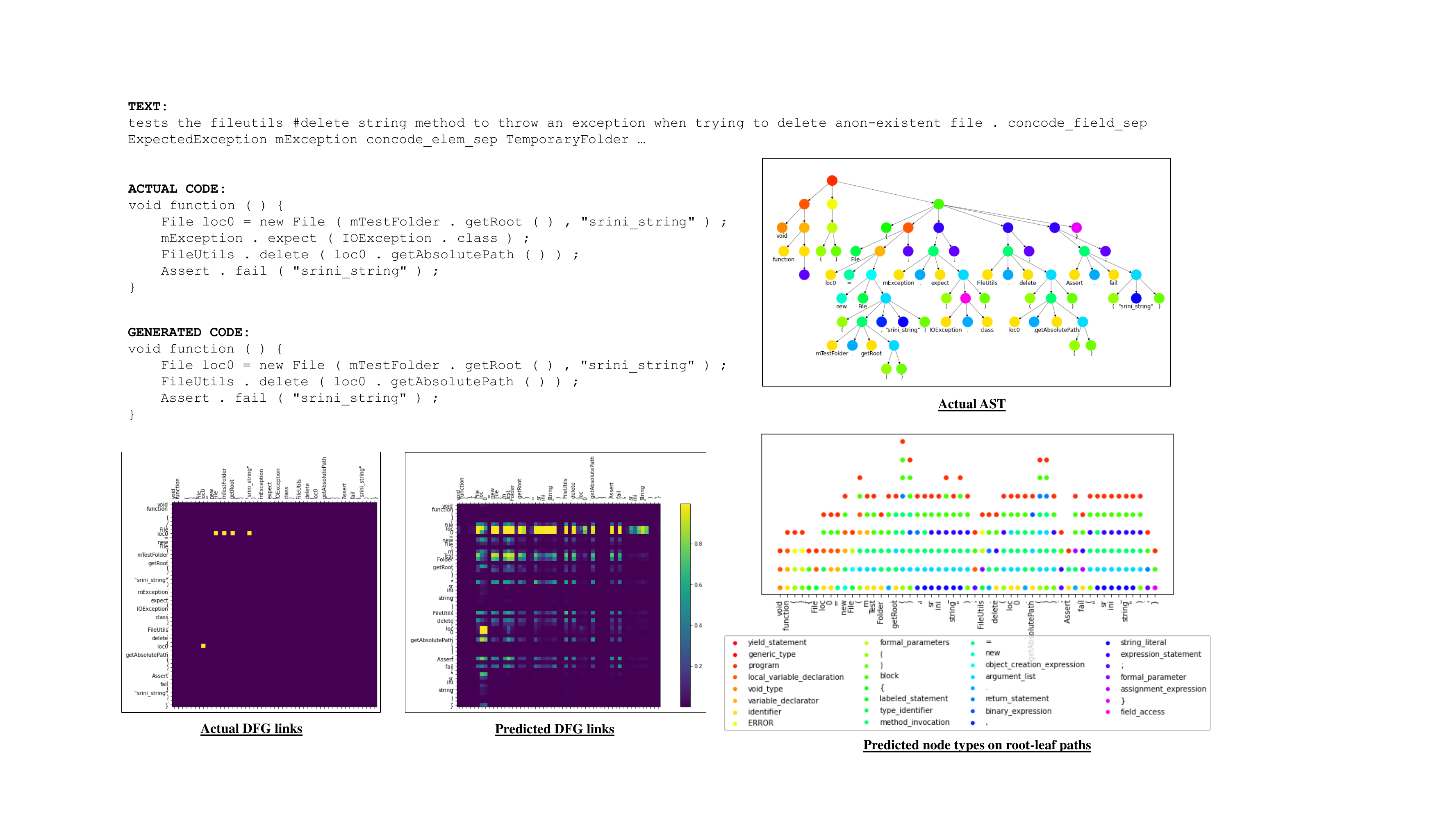}
    \caption{An example from the concode dataset with BLEU=78.85.}
    \label{fig:cs1}
\end{figure*}

\begin{figure*}[hbt!]
    \centering
    \includegraphics[scale=0.3, trim=0 0 10cm 1.2cm]{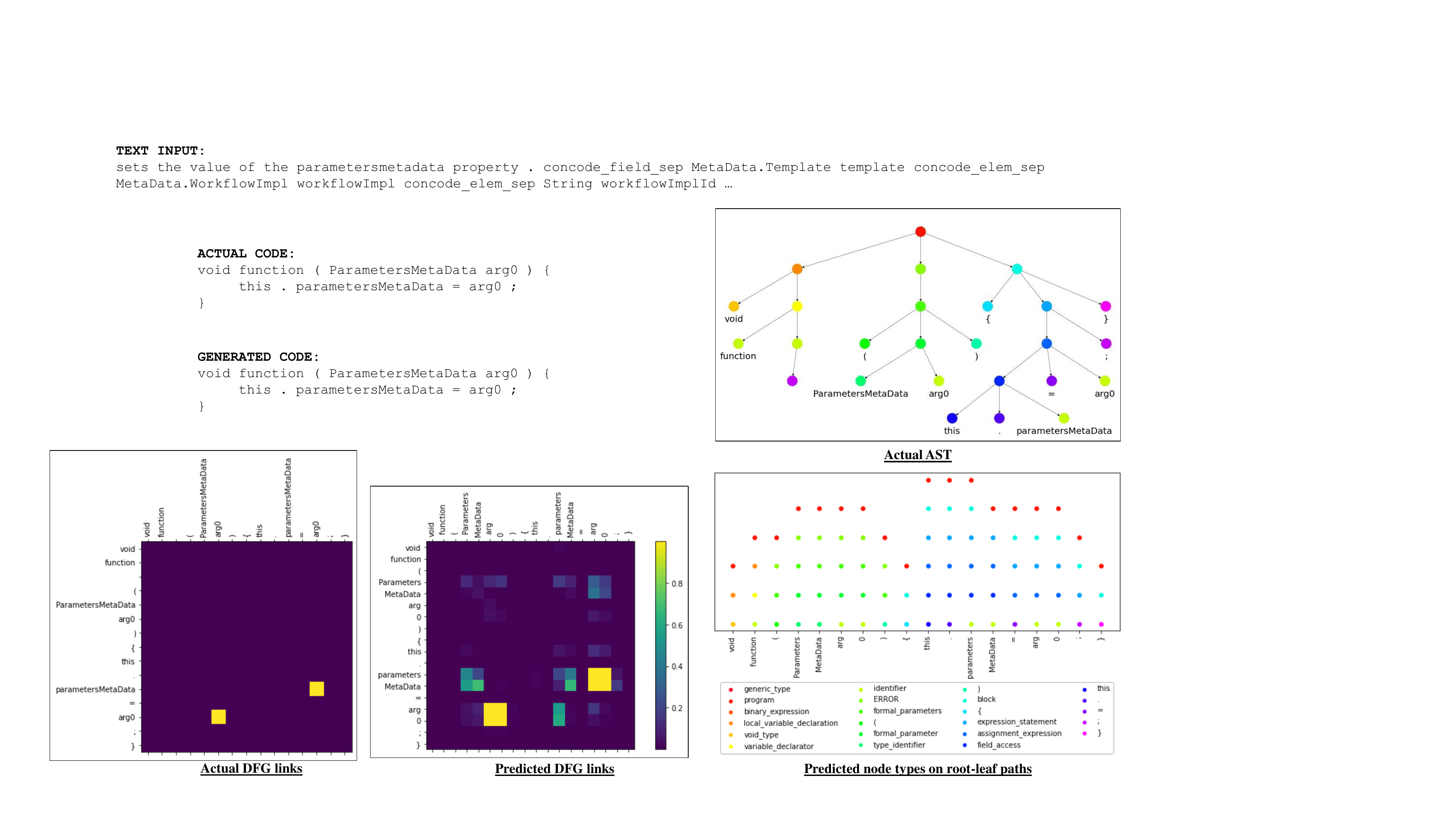}
    \caption{An example from the concode dataset with BLEU=100.}
    \label{fig:cs2}
\end{figure*}

\begin{figure*}[hbt!]
    \centering
    \includegraphics[scale=0.31, trim=3.5cm 0cm 9cm 5cm]{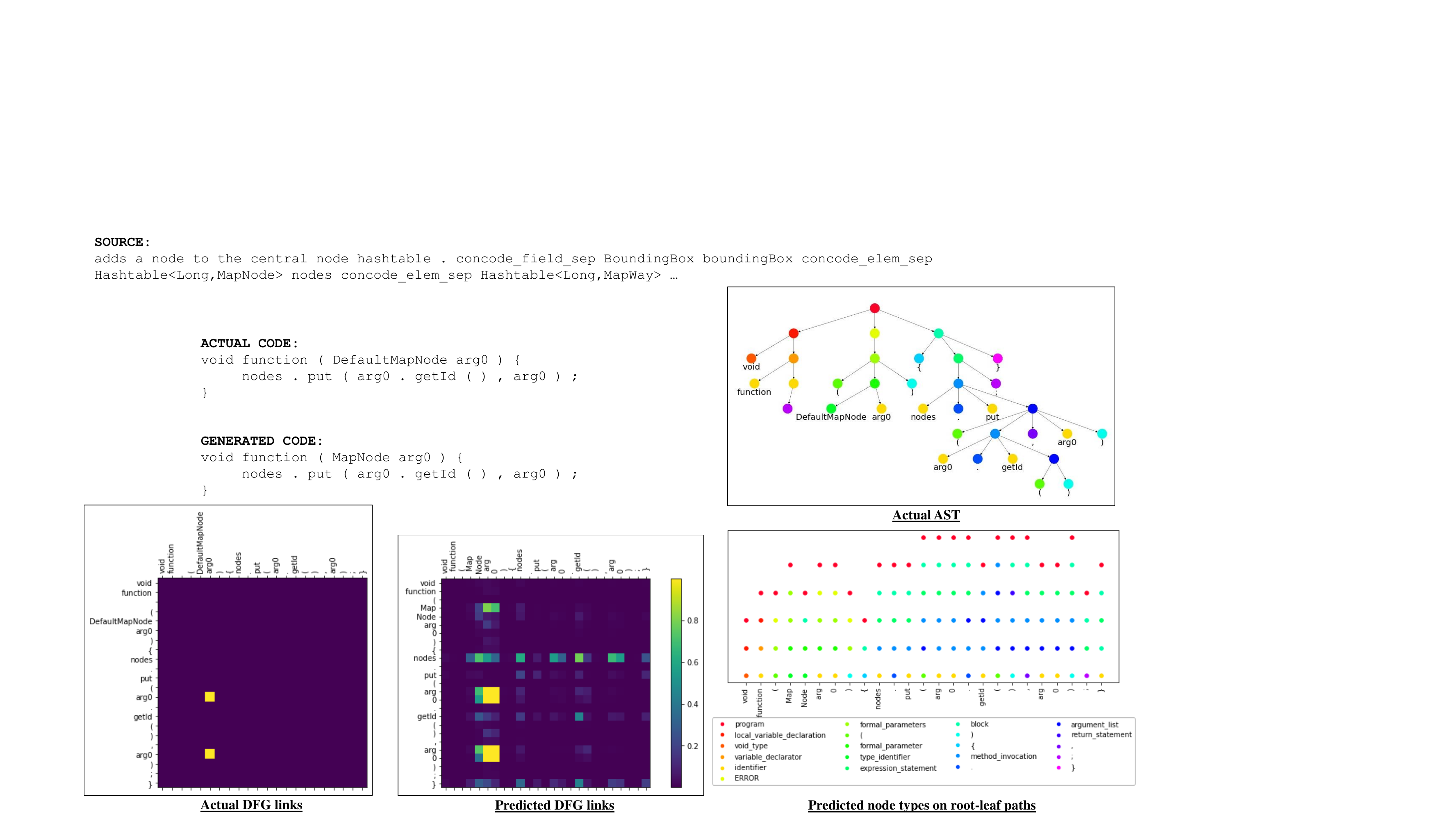}
    \caption{An example from the concode dataset with BLEU=87.25.}
    \label{fig:cs3}
\end{figure*}

In this section, we illustrate a few examples of text-to-code generation along with the predicted DFG links and AST paths (see Figures \ref{fig:cs1}-\ref{fig:cs3}). The DFG predictions are visualized as a matrix where the $ij^{th}$ cell denotes the probability of data flow from $j^{th}$ to $i^{th}$ token. To visualize predicted AST paths, for each predicted token, we indicate the predicted node types on the path starting from the root (top) to the leaf (bottom) containing this token, vertically using colored discs. 

\end{document}